\documentclass[10pt,twocolumn,letterpaper]{article}

\usepackage[pagenumbers]{cvpr} 

\usepackage[table,dvipsnames,svgnames]{xcolor}

\usepackage{multirow}
\usepackage{amsmath,amssymb}
\usepackage{xcolor}
\usepackage{enumitem}
\usepackage{booktabs}
\usepackage{graphicx}
\usepackage{amssymb}
\usepackage{booktabs}
\usepackage{tabularx}
\usepackage{adjustbox}

\def\etal{\textit{et al.}~}
\def\ie{\textit{i.e.,}~}
\def\eg{\textit{e.g.,}~}
\usepackage{xspace}
\makeatletter
\DeclareRobustCommand\onedot{\futurelet\@let@token\@onedot}
\def\@onedot{\ifx\@let@token.\else.\null\fi\xspace}

\usepackage{array}
\usepackage{helvet} 
\usepackage{microtype}
\newcolumntype{C}[1]{>{\centering}m{#1}}
\newcolumntype{M}[1]{>{\centering\let\newline\\\arraybackslash\hspace{-2pt}}m{#1}}

\definecolor{cvprblue}{rgb}{0.21,0.49,0.74}
\usepackage[pagebackref,breaklinks,colorlinks,allcolors=cvprblue]{hyperref}

\usepackage[utf8]{inputenc} 
\usepackage[T1]{fontenc}    
\usepackage{url}            
\usepackage{amsfonts}       
\usepackage{nicefrac}       
\usepackage{microtype}      
\usepackage{inconsolata}
\usepackage{caption}
\usepackage{amsmath}
\usepackage{amssymb}
\usepackage{mathtools}
\usepackage{enumitem}
\usepackage{makecell} 
\usepackage[usestackEOL]{stackengine}
\usepackage{graphicx}
\usepackage{capt-of}

\usepackage[most]{tcolorbox}

\usepackage{tikz}
\usetikzlibrary{shadows}
\usepackage{float}
\usepackage{subcaption}
\usepackage{xspace}
\usepackage{svg}
\usepackage{import}
\usepackage{animate}
\usepackage{arydshln}
\usepackage{colortbl}
\usepackage{pifont}
\usepackage{alltt}
\usepackage{algpseudocode}

\usepackage{textcomp}  

\definecolor{carolinablue}{rgb}{0.6, 0.73, 0.89}
\definecolor{mildgreen}{rgb}{0.85, 0.98, 0.80}
\definecolor{beautycolor}{rgb}{0.91, 0.75, 0.96} 
\definecolor{fallacycolor}{rgb}{0.85, 0.95, 1}
\definecolor{gendercolor}{rgb}{1, 0.85, 0.85}
\definecolor{brightyellow}{RGB}{255, 255, 100}
\definecolor{boxcolor}{RGB}{51,51,153}
\definecolor{lightgreen}{rgb}{0.56, 0.93, 0.56}
\definecolor{citeblue}{HTML}{0064E0}

\newcommand{\thickhline}{\noalign{\hrule height 1pt}}

\definecolor{titlegray}{rgb}{0.4, 0.4, 0.4} 
\definecolor{contentgray}{rgb}{0.95, 0.95, 0.95} 
\definecolor{deepblue}{RGB}{0, 0, 139}
\definecolor{carolinablue}{rgb}{0.6, 0.73, 0.89}
\definecolor{mildgreen}{rgb}{0.85, 0.98, 0.80}
\definecolor{beautycolor}{rgb}{0.91, 0.75, 0.96} 
\definecolor{fallacycolor}{rgb}{0.85, 0.95, 1}
\definecolor{gendercolor}{rgb}{1, 0.85, 0.85}
\definecolor{brightyellow}{RGB}{255, 255, 100}
\definecolor{boxcolor}{RGB}{51,51,153}
\definecolor{lightgreen}{rgb}{0.56, 0.93, 0.56}
\definecolor{citeblue}{HTML}{0064E0}

\definecolor{questionbg}{RGB}{240, 248, 255} 
\definecolor{answerbg2}{RGB}{255, 241, 204} 
\definecolor{answerbg}{RGB}{245, 255, 250} 
\definecolor{bordercolor}{RGB}{100, 149, 237} 
\definecolor{titlecolor}{RGB}{25, 25, 112} 
\definecolor{bordercoloryellow}{RGB}{255, 192, 0}
\definecolor{bordercolorgreen}{RGB}{0, 176, 80}
\definecolor{bordercolorblue}{RGB}{0, 112, 192}

\newtcolorbox{questionbanner}{
  colback=blue!10!white,    
  colframe=blue!80!black,   
  width=\textwidth,
  arc=4mm,                  
  boxrule=1pt,              
  fonttitle=\bfseries,
  title=Question,
}
\newtcolorbox{promptbox}[1]{
  colback=contentgray,      
  colframe=titlegray,       
  colbacktitle=titlegray,   
  coltitle=white,           
  title={#1}, 
  arc=4mm,                  
  rounded corners=northwest, 
  rounded corners=northeast, 
  boxrule=1pt,              
  fonttitle=\bfseries,      
}

\newtcolorbox{keywordbox}[1]{
  colback=white,      
  colframe=bordercolor,       
  colbacktitle=bordercolor,   
  coltitle=white,           
  title={#1}, 
  arc=4mm,                  
  rounded corners=northwest, 
  rounded corners=northeast, 
  boxrule=1pt,              
  fonttitle=\bfseries,      
}

\definecolor{questionbg}{RGB}{240, 248, 255}  
\definecolor{answerbg2}{RGB}{255, 241, 204}   
\definecolor{answerbg}{RGB}{245, 255, 250}   
\definecolor{bordercolor}{RGB}{100, 149, 237} 
\definecolor{titlecolor}{RGB}{25, 25, 112}    
\definecolor{bordercoloryellow}{RGB}{255, 192, 0}
\definecolor{bordercolorgreen}{RGB}{0, 176, 80}
\definecolor{bordercolorblue}{RGB}{0, 112, 192}

\newtcolorbox{vqaexample}[2][]{
    enhanced,
    breakable,
    colback=white,
    colframe=bordercolor,
    boxrule=1.5pt,
    arc=4pt,
    outer arc=4pt,
    left=8pt,
    right=8pt,
    top=8pt,
    bottom=8pt,
    drop shadow={shadow xshift=0.5mm, shadow yshift=-0.5mm, opacity=0.3},
    overlay={
        \node[
            anchor=north east,
            xshift=-3pt,
            yshift=-3pt,
            fill=bordercolor!80,
            text=white,
            font=\bfseries,
            rounded corners=2pt,
            inner sep=4pt,
            minimum height=1.2em,
            align=center
        ] at (frame.north east) {#2};
    },
    #1
}

\def\paperID{15958} 
\def\confName{CVPR}
\def\confYear{2026}

\title{Learning When to Look: A Disentangled Curriculum for Strategic Perception in Multimodal Reasoning}

\author{
Siqi Yang\textsuperscript{\rm 1}\footnotemark[1],
Zilve Gao\textsuperscript{\rm 2}\thanks{Equal contribution. Work done when Zilve Gao worked as an intern with Meituan.},
Haibo Qiu\textsuperscript{\rm 1}, 
Fanfan Liu\textsuperscript{\rm 1},
Peng Shi\textsuperscript{\rm 1},
Zhixiong Zeng\textsuperscript{\rm 1}, \\
Qingmin Liao\textsuperscript{\rm 2},
Lin Ma\textsuperscript{\rm 1}\footnotemark[2]\thanks{Corresponding author.} \\
    \textsuperscript{\rm 1}Meituan,
    \textsuperscript{\rm 2}Tsinghua University \\
    \tt \small{siqi.yang@uq.net.au, gaozl24@mails.tsinghua.edu.cn} \\
    \tt \small{liaoqm@tsinghua.edu.cn, forest.linma@gmail.com}
}
\begin{document}
\maketitle

\begin{abstract}
Multimodal Large Language Models (MLLMs) demonstrate significant potential but remain brittle in complex, long-chain visual reasoning tasks. A critical failure mode is ``visual forgetting'', where models progressively lose visual grounding as reasoning extends, a phenomenon aptly described as ``think longer, see less''. We posit this failure stems from current training paradigms prematurely entangling two distinct cognitive skills: (1) abstract logical reasoning (``how-to-think'') and (2) strategic visual perception (``when-to-look''). This creates a foundational cold-start deficiency--—weakening abstract reasoning--—and a strategic perception deficit, as models lack a policy for when to perceive. In this paper, we propose a novel curriculum-based framework to disentangle these skills. First, we introduce a disentangled Supervised Fine-Tuning (SFT) curriculum that builds a robust abstract reasoning backbone on text-only data before anchoring it to vision with a novel Perception-Grounded Chain-of-Thought (PG-CoT) paradigm. Second, we resolve the strategic perception deficit by formulating timing as a reinforcement learning problem. We design a Pivotal Perception Reward that teaches the model when to look by coupling perceptual actions to linguistic markers of cognitive uncertainty (e.g., ``wait'', ``verify''), thereby learning an autonomous grounding policy. Our contributions include the formalization of these two deficiencies and the development of a principled, two-stage framework to address them, transforming the model from a heuristic-driven observer to a strategic, grounded reasoner. \textbf{Code}: \url{https://github.com/gaozilve-max/learning-when-to-look}.
\end{abstract}

\begin{figure*}[htbp]
  \centering
    \includegraphics[width=\textwidth]{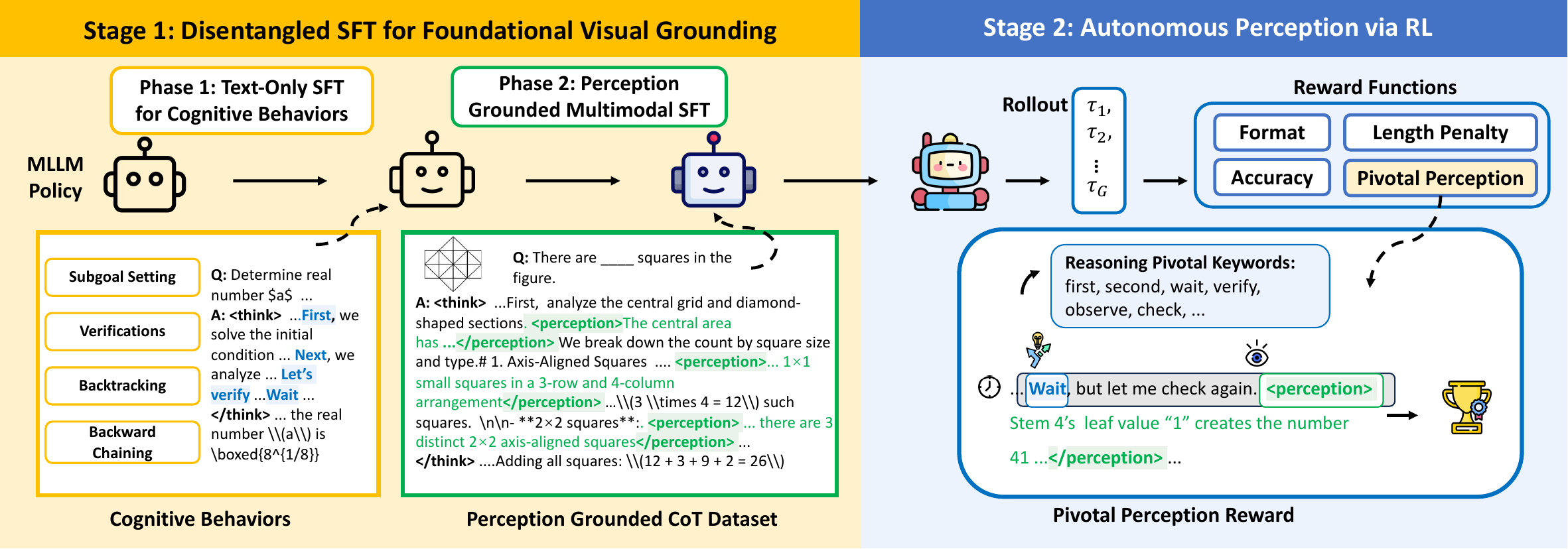}
  \caption{Overview of our curriculum-based framework for strategic perception in multimodal reasoning. Stage 1 (left) employs a disentangled SFT curriculum: Phase 1 conducts text-only SFT to build abstract reasoning capabilities (cognitive warm-up), and Phase 2 anchors these behaviors to visual evidence via Perception-Grounded Chain-of-Thought (PG-CoT) data. Stage 2 (right) learns autonomous perception timing through RL using a composite reward that includes our novel Pivotal Perception Reward, which couples perception actions to linguistic markers of cognitive uncertainty (e.g., ``wait'', ``verify'', ``first'').}
  \label{fig:overview}
\end{figure*}

\section{Introduction}
Multimodal Large Language Models (MLLMs) represent a significant step toward artificial general intelligence, enabling systems to jointly process and reason over visual and textual modalities in complex, real-world scenarios~\cite{comanici2025gemini,openai_o3_o4mini_2025,qwen3vl,guo2025seed1,team2025kimi}. 
The remarkable success of Large Language Models (LLMs) in reasoning tasks, demonstrated through chain-of-thought (CoT) prompting~\cite{wei2022chain,kojima2022large} and more recently through reinforcement learning with veifiable rewards (RLVR) approaches~\cite{shao2024deepseekmath,guo2025deepseek}, has naturally motivated the extension of reasoning capabilities to the multimodal domain.

Despite recent progress, MLLMs exhibit notable weaknesses in long-chain visual reasoning tasks~\cite{meng2025mm,zhou2025r1}. 
A critical issue arises in the form of the shortcut phenomenon: models frequently bypass visual analysis, relying instead on linguistic priors to infer answers~\cite{xia2025visionary,xiao2025advancing}. This leads to pronounced \textbf{visual forgetting}, succinctly captured by the notion ``think longer, see less''~\cite{tian2025more,wan2025srpo,sun2025mitigating}. As reasoning chains lengthen, models increasingly lose visual grounding, ultimately defaulting to degenerate solutions.
We argue that this brittleness arises from a failure in current training paradigms to disentangle two essential cognitive skills: (1) abstract logical reasoning (``how-to-think'') and (2) strategic visual perception (``when-to-look''). Standard reinforcement learning frameworks optimize for answer accuracy but do not explicitly penalize visual disengagement. As a result, models tend to converge to solutions where minimal visual processing is sufficient to satisfy the reward signal.

To mitigate these issues, several approaches have been proposed.
Global captioning methods, such as Visionary-R1~\cite{xia2025visionary} and Perception-R1~\cite{xiao2025advancing}, prepend image descriptions, providing superficial context but failing to enforce fine-grained reasoning
Reflection-based approaches~\cite{yang2025look,wan2025srpo,chung2025don,sun2025mitigating} interleave implicit re-focusing steps, but these operate post-hoc and cannot prevent initial visual disengagement.
Visual anchor based methods~\cite{sarch2025grounded,ni2025point,tian2025more} manually annotate visual evidence (e.g., bounding boxes) into reasoning chains, achieving fine-grained grounding at the expense of costly annotation and inflexible schemas.

However, these strategies share two fundamental limitations.
First, they prematurely entangle the training of abstract reasoning and visual grounding, which we term the \textbf{foundational cold-start deficiency}. Multimodal supervised fine-tuning (SFT) often uses data that is ``easier than text-only sets'', resulting in MLLMs with weaker textual reasoning than their LLM-only counterparts. By forcing models to learn ``how-to-think'' and ``what-to-see'' simultaneously, these approaches fail to establish a robust abstract ``System 2'' backbone for reasoning. This deficiency, rooted in the SFT stage, creates a cold-start problem that subsequent RL cannot fully overcome. Prior work shows that RL can only refine or compose behaviors already latent in the base model's distribution~\cite{gandhi2025cognitive,zeng2025simplerl}. For example, Gandhi et al.~\cite{gandhi2025cognitive} identified key cognitive behaviors in text-only settings—such as sub-goal setting, backtracking, and verification—that are prerequisites for RL-driven self-improvement. Models lacking these behaviors must first be bootstrapped via SFT on curated reasoning traces.

Second, existing methods do not learn \textbf{when to look}; instead, they impose perception through fixed heuristics, such as global captions at the outset~\cite{xia2025visionary,xiao2025advancing} or visual anchors at random steps~\cite{sarch2025grounded}.
This creates a fundamental limitation: the model learns \textit{to} perceive but not \textit{when} to perceive. Constrained to imitate predetermined grounding patterns, it cannot adapt to its own emerging cognitive state or question-specific demands. The result is either inefficient over-perception or critical under-perception, with no mechanism to acquire strategic timing.

In this paper, we propose a novel curriculum-based framework that disentangles the learning of these two skills.
To remedy the foundational cold-start deficiency, we introduce a \textbf{disentangled SFT curriculum}. First, we conduct text-only SFT on high-difficulty reasoning data, forcing the model to acquire robust abstract logical templates and cognitive behaviors without visual confounding. We then anchor vision through a novel Perception-Grounded Chain-of-Thought (PG-CoT) paradigm, where a teacher model annotates fine-grained, step-relevant perception segments into existing reasoning traces. This two-phase curriculum builds a strong textual reasoning backbone first, then teaches it to bind to visual evidence, establishing the robust antecedent skills necessary for strategic self-improvement.

To resolve the strategic perception deficit, we formulate perception timing as a reinforcement learning problem. While SFT teaches the \textit{structure} of grounded reasoning, RL learns \textit{when} to invoke it. Our key innovation is a {Pivotal Perception Reward} that evaluates perception placement against linguistic markers of cognitive uncertainty. We operationalize the hypothesis that strategic perception should be triggered by metacognitive signals (e.g., ``wait, I was mistaken''), structural transitions (e.g., ``however''), and explicit re-examination cues (e.g., ``verify this''). Our reward analyzes perception actions at the sentence level, measuring their coupling to these pivotal keywords while regularizing against heuristic overfitting. This compels the model to develop a nuanced, context-aware policy that binds visual grounding to genuine cognitive uncertainty.

Our primary contributions are:
\begin{itemize}
\item  We identify and formalize foundational reasoning weakness in MLLM training, attributing it to entangled SFT objectives. We propose a {disentangled SFT curriculum} that sequentially builds abstract reasoning priors before anchoring them to vision, establishing robust antecedent capabilities for RL.
\item  We introduce a Perception-Grounded CoT data paradigm, preserves reasoning fidelity while teaching perceptual discipline through fine-grained, step-relevant perception segments, serving as the enabling bridge between foundational reasoning and strategic grounding.
\item  We resolve the strategic perception deficit by designing the first RL framework that learns autonomous perception timing, featuring a pivotal perception reward that couples visual grounding to linguistic markers of cognitive uncertainty with selectivity regularization.
    
\end{itemize}

\section{Related Works}
\noindent
\textbf{Multimodal Reasoning.} Deepseek~\cite{guo2025deepseek} introduced the GRPO training approach to equip large language models with chain-of-thought reasoning abilities, enabling them to think before answering. Building on this, many subsequent works have focused on enhancing GRPO and applying it to MLLMs to improve their reasoning skills. 
VL-rethinker~\cite{wang2025vl} proposes Selective Sample Replay to address the diminishing advantages of GRPO, and Forced Rethinking to encourage self-reflection in vision-language models during RL training. WE-MATH 2.0~\cite{qiao2025wemath2} enhances MLLMs' visual mathematical reasoning through a five-level knowledge system, 3D difficulty-modeled datasets, and a two-stage progressive RL framework with dynamic scheduling. ThinkLite-VL~\cite{wang2025sota} first collects 70k samples, uses Monte Carlo Tree Search (MCTS) to select challenging ones via iterative reasoning, and then applies Reinforcement Fine-Tuning (RFT) on the filtered 11k samples without knowledge distillation. ARES~\cite{chen2025ares} leverages window entropy as exploration triggers and difficulty-aware rewards to dynamically adjust reasoning effort for multimodal large reasoning models via adaptive training. However, most RL-based MLLM research has mainly focused on boosting models' textual reasoning abilities, while the enhancement of perceptual capabilities remains largely overlooked.

\noindent\textbf{Perception Enhancement in Multimodal Reasoning.}
The pervasive shortcut phenomenon, where models over-rely on linguistic priors and suffer from visual forgetting~\cite{xia2025visionary,tian2025more}, has motivated diverse grounding strategies. \textit{Global captioning} methods (e.g., Visionary-R1~\cite{xia2025visionary}, Perception-R1~\cite{xiao2025advancing}) prepend holistic image descriptions to provide contextual priming, yet they fail to enforce fine-grained, step-relevant grounding and cannot prevent mid-reasoning visual disengagement. Reflection-based approaches~\cite{yang2025look,wan2025srpo,sun2025mitigating} interleave post-hoc re-examination tokens (e.g., \texttt{back}) to encourage visual revisitation, but these operate reactively and cannot preempt initial disengagement. More structured are \textit{visual anchor} methods: Point-RFT~\cite{ni2025point} and Visual CoT~\cite{shao2024visual} annotate each reasoning step with explicit bounding boxes or region pointers, achieving precise alignment at the cost of expensive manual labeling and inflexible, predetermined grounding schemas. PeBR~\cite{chen2025perception} and GeoPQA~\cite{chen2025geopqa} adopt two-stage RL pipelines that first optimize perception via fine-grained visual rewards (e.g., FG-CLIP alignment) before reasoning training. Crucially, these methods conflate perception enhancement with perception scheduling: they teach models how to ground but not when to strategically invoke visual analysis. Consequently, they impose perception through fixed heuristics—global captions at onset, random anchors, or reflection tokens at predetermined steps—preventing the model from acquiring a dynamic, context-aware perception policy. In contrast, our RL framework learns autonomous timing by coupling perception actions to linguistic markers of cognitive uncertainty, enabling adaptive grounding that responds to the model's internal reasoning state.


\section{Methodology}
\label{sec:method}

We propose a curriculum-based framework that disentangles the acquisition of abstract reasoning from visual grounding. Our approach comprises two principled stages: (1) a disentangled supervised fine-tuning curriculum that establishes a robust reasoning foundation, and (2) a reinforcement learning phase that learns strategic perception timing. Figure~\ref{fig:overview} provides an overview of our framework.

\subsection{Disentangled SFT for Foundational Visual Grounding}
\label{sec:stage1}
The shortcut phenomenon fundamentally stems from conflating two distinct learning objectives: acquiring abstract reasoning patterns and grounding them in visual evidence. We address this through a two-step curriculum that sequentially develops these capabilities.

\subsubsection{Phase 1: Cognitive Warm-up via Text-Only SFT}

To build a strong foundation for abstract reasoning, we first train the vision-language model without the confounding influence of visual inputs. 
This approach allows the model to learn abstract logical templates, problem decomposition, and other cognitive behaviors (e.g., verification, subgoal setting, backtracking) purely from high-difficulty, text-only reasoning data.

Formally, let $\mathcal{D}_{\text{text}} = {(x_i, y_i)}_{i=1}^{N_{text}}$ denote a text-only reasoning dataset where $x_i$ is the query and $y_i = [y_{i,1}, \dots, y_{i,T_i}]$ is the tokenized chain-of-thought rationale of length $T_i$. We optimize the standard auto-regressive objective:
\begin{equation}
\mathcal{L}_{\text{SFT-text}} = -\mathbb{E}_{(x, y) \sim \mathcal{D}_{\text{text}}} \left[ \sum_{t=1}^{T} \log \pi_{\theta}(y_t \mid x, y_{<t}) \right] \textrm{ ,}
\label{eq:sft_text}
\end{equation}
where $\pi_\theta$ is the vision-language model (e.g., Qwen2.5-VL~\cite{bai2025qwen2}) with vision encoder frozen, and $y_{<t}$ denotes previously generated tokens. This stage yields policy $\pi_{\theta_{T}}$ with strong textual reasoning priors.

\begin{figure*}[t]
    \centering
        \centering
        \includegraphics[width=1.0\linewidth]{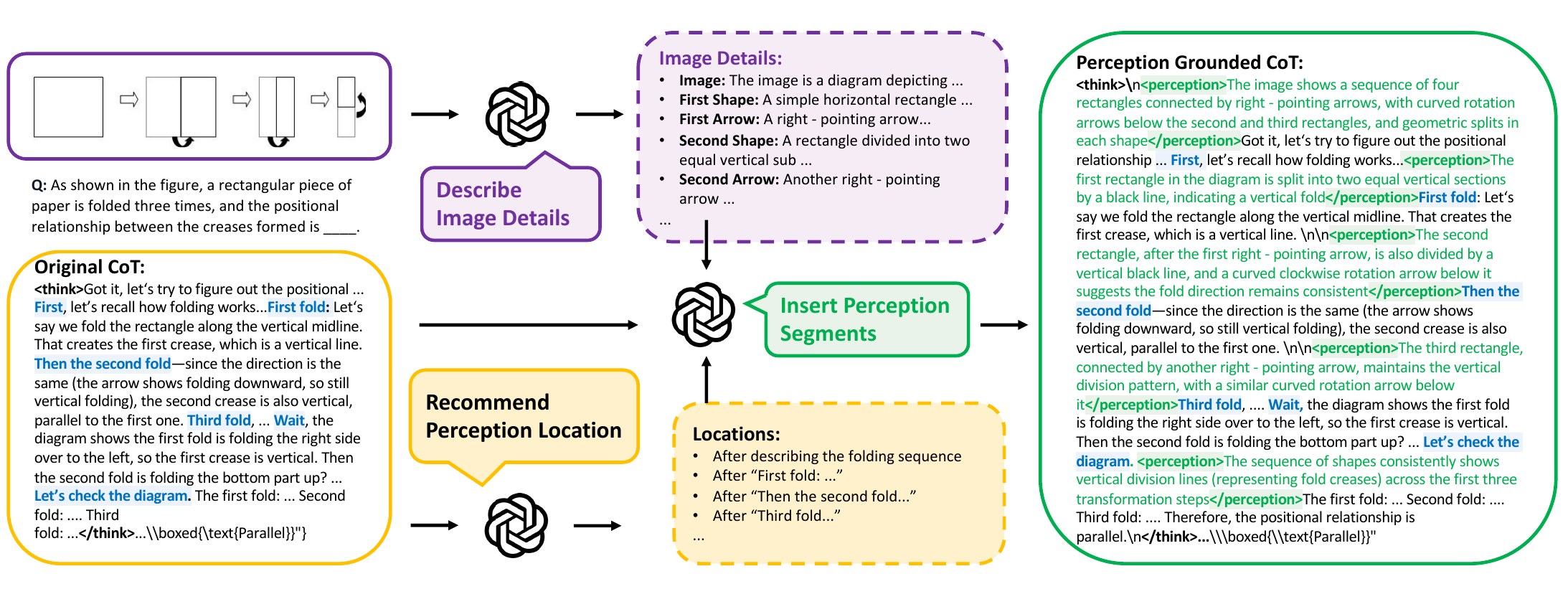}
        \caption{Data generation pipeline for Perception-Grounded Chain-of-Thought (PG-CoT).Using a teacher MLLM (e.g., GPT-4o), we transform standard CoT reasoning traces by identifying logical breakpoints where visual verification is required (marked as ``Recommend Perception Location'') and inserting fine-grained observation segments. }
        \label{fig:data_pipeline}
\end{figure*}

\subsubsection{Phase 2: Anchoring Logic with Perception-Grounded SFT}
\label{sec:stage2}
Building upon $\pi_{\theta_{T}}$, we anchor abstract reasoning to visual evidence through a novel Perception-Grounded Chain-of-Thought (PG-CoT) paradigm. This stage teaches the model to interleave perceptual verification with reasoning steps.

\noindent
\textbf{PG-CoT Data Construction.} We transform existing multimodal reasoning datasets (e.g., ARES~\cite{chen2025ares}) into PG-CoT format using a teacher MLLM (e.g., GPT-4o or Seed1.5-VL). 
As illustrated in Figure~\ref{fig:data_pipeline}, the data generation pipeline operates in three steps: (1) analyze the original CoT to identify logical breakpoints where visual evidence is required, (2) generate fine-grained observations anchored to specific visual elements, and (3) insert perception segments while preserving reasoning fidelity.
Given an image $I$, question $x$, and original CoT, $y$, the teacher identifies logical breakpoints where visual evidence is required, then inserts explicit perception segments:

\newcommand{\stag}[1]{{\small\texttt{#1}}}

\begin{align*}
\begin{split}
\hat{y} = [\stag{<think>}, \dots, \underbrace{\stag{<perception>} p_1 \stag{</perception>}}_{\text{perception segment 1}}, \dots, \\
\qquad \underbrace{\stag{<perception>} {p}_2 \stag{</perception>}}_{\text{perception segment 2}}, \dots, \stag{</think>}] \textrm{ ,}
\end{split}
\end{align*}
where ${p}$ is a fine-grained observation derived solely from $I$, adhering to three constraints: (i) ${p}$ must be factually verifiable from $I$; (ii) ${p}$ must be necessary for $y_t$; (iii) the original logical structure of $y$ must remain unaltered. This annotation style preserves reasoning fidelity while teaching perceptual discipline. 

\noindent
\textbf{Multimodal SFT Objective.}
Crucially, in contrast to the image caption based methods~\cite{xia2025visionary,yu2025perception}, the content of the ``\texttt{<perception>}'' tags are not global descriptions, but fine-grained observations that are directly relevant to the reasoning step at hand. 
We fine-tune $\pi_{\theta_{T}}$ on the constructed PG-CoT dataset $\mathcal{D}_{\text{mm}} = {(I_i, x_i, \hat{y}_i)}$:

\begin{equation}
\mathcal{L}_{\text{SFT-mm}} = -\mathbb{E}_{(I, x, \hat{y}) \sim \mathcal{D}_{\text{mm}}} \left[ \sum_{t=1}^{\hat{T}} \log \pi_\theta(\hat{y}_t \mid I, x, \hat{y}_{<t}) \right],
\label{eq:sft_mm}
\end{equation}

where the vision encoder remains frozen. This yields policy $\pi_{\theta_{M}}$ capable of generating structured reasoning with interleaved perception. However, $\pi_{\theta_{M}}$ merely imitates the fixed perception schedule of the teacher, lacking strategic autonomy.

\subsection{Autonomous Perception via Reinforcement Learning}
\label{sec:stage2}

A fundamental limitation of prior methods~\cite{sarch2025grounded} addressing visual forgetting is their reliance on randomly inserted perception segments—either through rigid heuristics or human-annotated grounding points. While effective at reintroducing visual evidence, these approaches impose a fixed perception schedule that cannot adapt to the model's emerging reasoning patterns or question-specific demands. This rigidity creates a critical gap: the model learns to perceive but not when to perceive, remaining a passive executor rather than an autonomous agent.

We close this gap by formulating perception timing as a strategic decision-making problem that must be learned. 
While supervised fine-tuning teaches the model to imitate the structure of grounded reasoning, it produces a rigid policy that cannot adaptively decide when visual grounding is necessary.
The key insight is to reward not just \textit{what} the model perceives, but \textit{when} it chooses to perceive.

\subsubsection{Reward Design for Cognitive-Perceptual Coupling}
Our composite reward function encourages both reasoning correctness and strategic perceptual grounding. The total reward is a weighted sum of four components:
\begin{equation}
R = \lambda_{\text{acc}} R_{\text{acc}} + \lambda_{\text{form}} R_{\text{form}} + \lambda_{\text{pivot}} R_{\text{pivot}} + \lambda_{\text{len}} R_{\text{len}} \textrm{ .}
\end{equation}

\noindent\textbf{Pivotal Perception Reward ($R_{\text{pivot}}$).} 
This reward component is designed to train the model's policy on when to perceive. 
We formulate the hypothesis that strategic perception should be triggered by linguistic markers of cognitive uncertainty. 
To this end, we curate a comprehensive lexicon of reasoning-pivotal keywords, $\mathcal{K}$, which includes: (i) metacognitive markers (e.g., ``wait'', ``I was mistaken''), (ii) structural transitions (e.g., ``first'', ``step 2'', ``however''), and (iii) visual re-examination cues (e.g., ``observe'', ``verify'', ``check again''). (Full list in the Supplementary Material).

Rather than employing token-level heuristics that disregard clausal boundaries, we analyze perception placement at the sentence level, where cognitive state transitions naturally manifest. 
For a given trajectory $\tau$, let $\mathcal{P}_\tau = \{p_1, \dots, p_m\}$ be the ordered set of $m$ perception segments (i.e., \texttt{<perception>} tokens) generated. 
For each perception action $p_i \in \mathcal{P}_\tau$, let $s(p_i)$ denote the sentence immediately preceding it.
We first define the set of keyword-coupled perceptions, $\mathcal{P}_{\text{coupled}}$, as the subset of $\mathcal{P}_\tau$ where the preceding sentence $s(p_i)$ contains a pivotal keyword from our lexicon:
\begin{equation}
    \mathcal{P}_{\text{coupled}} = \{ p_i \in \mathcal{P}_\tau \mid \exists \kappa \in \mathcal{K}, \kappa \in s(p_i) \} \textrm{ .}
\end{equation}
Let $m_{\text{coupled}} = |\mathcal{P}_{\text{coupled}}|$ be the count of such coupled actions.
The pivotal score, $S_{\text{pivot}}(\tau)$, is then defined as the proportion of perception actions that are anchored to a preceding pivotal keyword:
\begin{equation}
S_{\text{pivot}}(\tau) = \begin{cases} \frac{m_{\text{coupled}}}{m}, & \text{if } m > 0 \\ 0,  & \text{if } m = 0 \end{cases} \textrm{ .}
\end{equation}
Rewarding $S_{\text{pivot}}(\tau)$ alone could lead to a degenerate policy that overfits to these keywords. To mitigate such heuristic overfitting, we introduce a selectivity regularizer. We define a maximum allowable coupling ratio, $\alpha_{\text{ratio}}$ (e.g., $\alpha_{\text{ratio}} = 0.5$), and a regularization coefficient, $\lambda_{\text{reg}}$.This regularizer is applied based on the excess coupling, $m_{\text{excess}}$, which quantifies the number of coupled perceptions beyond the allowed ratio:
$m_{\text{excess}} = \max\left(0, m_{\text{coupled}} - \alpha_{\text{ratio}} \cdot m\right)$.
The final pivotal perception reward, $R_{\text{pivot}}$, is the regularized score, clipped at zero:
\begin{equation}
R_{\text{pivot}}(\tau) = \max\left(0, S_{\text{pivot}}(\tau) - \lambda_{\text{reg}} \cdot m_{\text{excess}}\right) \textrm{ .}
\end{equation}
This dual mechanism simultaneously rewards keyword-coupled perception while applying a regularization term to discourage over-coupling. It compels the model to develop a more nuanced, context-aware policy that binds visual grounding to genuine cognitive uncertainty rather than engaging in superficial pattern matching.

\noindent
\textbf{Accuracy Reward ($R_{\text{acc}}$):} This is the primary task reward. $R_{\text{acc}} = 1$ if the final generated answer is correct, and 0 otherwise.

\noindent
\textbf{Format Reward ($R_{\text{form}}$):} A binary reward $R_{\text{form}} = 1$ if the generated output correctly follows the nested XML structure (e.g., all ``\texttt{<perception>}'' tags are closed and are within a ``\texttt{<think>}'' block). This penalizes malformed generation.
\noindent
\textbf{Length Reward ($R_{\text{len}}$):} To penalize overly verbose correct answers, we add a penalty for generations that exceed a gold length $L_{\text{gold}}$ (e.g., 2048 tokens). 
This reward is 0 if $R_{\text{acc}} = 0$. Otherwise, for a trajectory of length $L(\tau)$, the penalty is applied to the excess length, $\Delta L = L(\tau) - L_{\text{gold}}$:
\begin{equation}
R_{\text{len}}(\tau) = \begin{cases} 1, & \text{if } \Delta L \le 0 \\ \max(0, \delta - \lambda_{\text{len}} \Delta L), & \text{if } \Delta L > 0 \end{cases} \textrm{ ,}
\end{equation}
where $\delta$ and $\lambda_{\text{len}}$ are hyperparameters controlling the penalty curve.

\subsubsection{Policy Optimization}
We optimize our policy using DAPO~\cite{yu2025dapo}, which stabilizes training with dynamic clipping and sample filtering. Given a query-answer pair $(q, a)$ sampled from the data pool $\mathcal{D}$, the behavior policy model $\pi_{\theta_\text{old}}$ generates a group of $G$ candidate trajectories $\{ \tau_i \}_{i=1}^G$. The objective is:
\begin{multline}
\mathcal{J}_{\text{RL}}(\theta) = 
\mathbb{E}_{(q,a)\sim \mathcal{D}, \{\tau_i\}_{i=1}^G\sim \pi_{\theta_\text{old}}(\cdot\mid q)}\\
\Bigg[\frac{1}{N} \sum_{i=1}^{G}\sum_{t=1}^{|\tau_i|}
\min \Big( r_{i,t}(\theta) \hat{A}_{i,t}, 
\text{clip} ( r_{i,t}(\theta), 1 - \varepsilon_{\text{l}}, 
1 + \varepsilon_{\text{h}} ) \hat{A}_{i,t} \Big) \Bigg]\\
\text{s.t.}\quad 0< |\{\tau_i\mid\texttt{is\_equal}(gt,\tau_i)\}|< G \textrm{ ,}
\label{eq:dapoloss_oversample_filter}
\end{multline}
where $N$ is $\sum_{i=1}^{G}|\tau_i|$ and the importance ratio and advantage are calculated as:
\begin{equation}
    r_{i,t}(\theta)=\frac{\pi_{\theta}(\tau_{i,t} \mid q, \tau_{i,<t})}{\pi_{\theta_{\text{old}}}(\tau_{i,t} \mid q,\tau_{i,<t})}, \hat{A}_{i,t} = \frac{R_i - \text{mean}(\{R_i\}_{i=1}^G)}{\text{std}(\{R_i\}_{i=1}^G)} \textrm{ .}
\label{eq:advantage_calculation}
\end{equation}

This decoupled optimization prevents catastrophic forgetting of visual features, while enabling sophisticated reasoning-perception coordination.

\begin{table*}[t] 
\centering
\caption{Performance comparison of various MLLMs on multimodal reasoning benchmarks. Within the 7B model group, the best results are highlighted in \textbf{bold}; the second best are \underline{underlined}.}
  \setlength{\tabcolsep}{2pt}
  \renewcommand{\arraystretch}{1.1}
  \small
  \setlength{\extrarowheight}{2pt}
\begin{tabular}{l|cccccc|c} 
\thickhline
\multirow{2}{*}{Method} & \multicolumn{7}{c}{Multimodal Reasoning Benchmarks} \\
\cmidrule(lr){2-8}
& MathVerse-V & MathVision & MathVista & DynaMath-W & WeMath & LogicVista & Avg. \\
\hline
\multicolumn{8}{l}{\textbf{Closed-Source MLLMs}} \\
\hline
GPT-4.1~\cite{openai_gpt4_1_2025} & 59.8 & 51.8 & 72.0 & 48.3 & 55.5 & 63.8 & 58.5\\
Gemini-2.5-Pro-Thinking~\cite{comanici2025gemini} & 81.2 & 55.3 & 83.8 & 57.1 & 78.0 & 75.2 & 71.8 \\
Claude-4-Sonnet~\cite{anthropic_claude_sonnet_4} & 66.1 & 54.6 & 70.4 & 46.9 & 63.0 & 64.4 & 60.9 \\
Seed-1.5-thinking~\cite{guo2025seed1} & 80.4 & 68.7 & 85.6 & 60.5 & 78.0 & 71.8 & 74.2 \\
\hline
\multicolumn{8}{l}{\textbf{Open-Source 7B MLLMs}} \\
\hline
Qwen2.5-VL~\cite{bai2025qwen2} & 42.9 & 25.1 & 68.2 & 21.2 & 36.2 & 45.0 & 39.8 \\
OpenVLThinker~\cite{deng2025openvlthinker} & 40.7 & 25.9 & 72.3 & 21.2 & 37.9 & 41.4 & 39.9 \\
MM-Eureka~\cite{meng2025mm} & 49.6 & 26.9 & 73.0 & 24.0 & 34.7 & 46.8 & 42.5 \\
MathBook~\cite{qiao2025we} & 45.2 & 28.0 & 73.0 & / & \underline{48.4} & / & / \\
MMR1-Math-v0~\cite{leng2025mmr1} & 45.1 & 30.2 & 71.0 & 25.2 & 33.2 & 50.8 & 42.6 \\
ThinkLite-VL~\cite{wang2025sota} & 45.3 & 32.9 & 75.1 & 22.0 & 26.5 & 40.7 & 40.4 \\
VLAA-Thinking~\cite{chen2025sft} & 48.2 & 26.4 & 68.0 & 22.4 & 29.2 & 48.5 & 40.5 \\
VL-Rethinker~\cite{wang2025vl} & 49.1 & 32.3 & 74.9 & 27.4 & 27.8 & 44.5 & 42.7 \\
Vision-G1~\cite{zha2025vision} & 50.0 & 31.3 & 76.1 & 27.2 & 45.1 & 50.2 & 46.7 \\
Look-Back~\cite{yang2025look} & 51.8 & 30.3 & 72.3 & / & / & / & / \\
PeBR-R1~\cite{chen2025perception} & / & 32.7 & \textbf{76.0} & / & / & / & / \\
VAPO-Thinker~\cite{tian2025more} & \underline{53.3} & 31.9 & \underline{75.6} & / & 43.6 & \underline{50.9} & / \\
Revisual-R1~\cite{chen2025advancing} & \textbf{53.6} & \textbf{44.7} & 73.1 & \underline{27.5} & 42.0 & \textbf{52.3} & \underline{48.9} \\
\hline
Ours & \underline{53.3} & \underline{44.2} & 73.4 & \textbf{30.3} & \textbf{54.4} & 49.9 & \textbf{50.9} \\
\thickhline
\end{tabular}
\label{tab:multimodal_performance_selected}
\end{table*}

\section{Experiments}

\subsection{Implementation Details}
\noindent
\textbf{Training Details.}
Experiments were conducted on eight NVIDIA H100 GPUs. 
During the SFT stage, we utilized the LLaMA-Factory framework~\cite{zheng2024llamafactory} and froze the vision tower to preserve its pre-trained visual perception capabilities, training for 5 epochs with a learning rate of $5e^{-5}$ to balance effectiveness and prevent overfitting. Building upon this, we further performed reinforcement learning using the EasyR1 framework~\cite{zheng2025easyr1}, enabling full-parameter fine-tuning and conducting training for 2 epochs. Additional training configurations are provided in the Supplementary.

\noindent
\textbf{Training Datasets.}
For the perception SFT stage, we constructed our training data from collected open-source resources, which include 120k multimodal question-answer pairs and 100k pure-text QA pairs. To enhance reasoning quality, we generated our perception COT data by inserting precise image descriptions into multimodal answers, using multi-turn generation and filtering with the Seed-1.5-thinking-vision-pro model~\cite{guo2025seed1}. This process ensured that the dataset contained rich perceptual cues to support complex visual reasoning. For the RL stage, we further trained our model on the Virl39K dataset~\cite{wang2025vl}, which provides a diverse set of visual reasoning tasks.

\noindent 
\textbf{Benchmarks.} We evaluated our model's mathematical reasoning capabilities on several multimodal reasoning benchmarks, including MathVista~\cite{lu2023mathvista}, MathVision~\cite{wang2024measuring}, WeMath~\cite{qiao2025we},  MathVerse~\cite{zhang2024mathverse}, DynaMath~\cite{zou2024dynamath} and LogicVista~\cite{xiao2024logicvista}. These benchmarks cover a broad range of mathematical problem types and visual reasoning scenarios.
We also evaluate our method on perception-centric benchmarks: HallusionBench~\cite{guan2024hallusionbench},  MME~\cite{fu2025mme}, MMStar~\cite{chen2024we}, MMBench~\cite{liu2024mmbench}, MMMU~\cite{yue2023mmmu} and MMVet~\cite{yu2023mm}.

\subsection{Evaluation}
We evaluate our method on both multimodal reasoning and perception benchmarks.

\subsection{Evaluation on Multimodal Reasoning Benchmarks}

We compare our method against a comprehensive set of leading closed-source and open-source MLLMs on six challenging multimodal reasoning benchmarks. As shown in Table~\ref{tab:multimodal_performance_selected}, our approach establishes a new state-of-the-art for open-source 7B models.
While powerful proprietary models like Seed-1.5-thinking (74.2 Avg.) and Gemini-2.5-Pro-Thinking (71.8 Avg.) set a high performance ceiling, our 7B model demonstrates a significant advancement in the open-source domain.
Our method demonstrates particular strength on benchmarks requiring strategic visual perception. On WeMath, which emphasizes word problems with crucial visual details, we achieve \textbf{54.4}, surpassing the prior best (MathBook, 48.4) by 6.0 points, attesting to the efficacy of our Pivotal Perception Reward in learning \textit{when} to perceive.
Similarly, on DynaMath-W that tests dynamic mathematical reasoning, we attain \textbf{30.3}, significantly exceeding Revisual-R1's 27.5. 
These gains validate that our disentangled curriculum and strategic timing mechanism effectively mitigate visual forgetting in reasoning-heavy tasks.
On MathVision and MathVerse-V, which focus on textbook-style geometry and visual mathematical reasoning, we achieve competitive scores of \textbf{44.2} and \textbf{53.3} respectively, approaching or matching the best reported results. 
The superior performance on perception-critical benchmarks (WeMath, DynaMath-W) substantiates our core hypothesis: decoupling reasoning foundation from strategic perception learning enables MLLMs to dynamically ground their thinking in visual evidence when necessary.

\subsection{Evaluation on Multimodal Perception Benchmarks}

We extend our evaluation to a comprehensive suite of perception-centric and reasoning benchmarks to validate that our framework achieves a superior balance between foundational vision and high-level logic. Table~\ref{tab:multimodal_perception} compares our method against representative baselines across diverse metrics, including the average mathematical reasoning score (Avg. Math) from Table~\ref{tab:multimodal_performance_selected} for holistic comparison.
A prevalent challenge in RL-tuned MLLMs is the degradation of visual feature representations—often termed ``visual forgetting''—as the model optimizes for textual reasoning. However, our method significantly outperforms the backbone model, Qwen2.5-VL, across nearly all evaluated dimensions. We observe substantial gains on integrated reasoning tasks such as MMMU (+6.5) and MMVet (+6.4), alongside robust improvements on foundational perception benchmarks like MME (+127) and HallusionBench (+3.7). This confirms that our disentangled SFT curriculum and Pivotal Perception Reward effectively preserve and sharpen the model's visual grounding capabilities, successfully averting the visual forgetting problem of standard reasoning-focused fine-tuning.

Furthermore, our model establishes a new state-of-the-art among 7B models on MMMU (59.0) and MMVet (73.5). Unlike pure recognition tasks, these benchmarks require the high-order integration of visual extraction with logical inference. This performance aligns with our core hypothesis: our model does not merely memorize static visual features but learns a dynamic policy for strategic visual extraction, enabling it to support complex, multi-step problem-solving.

It is crucial to highlight the trade-off observed in competing approaches. While models focused explicitly on perception enhancement, such as ThinkLite-VL~\cite{wang2025sota}, achieve marginally higher scores on perception-centric recognition benchmarks (\eg MME: 2378 vs. ours 2307), this specialization comes at a severe cost to reasoning capabilities. As evidenced in Table~\ref{tab:multimodal_performance_selected}, ThinkLite-VL significantly underperforms on mathematical reasoning benchmarks (Avg.: 40.4 vs. Ours 50.9) and lags behind on integrated tasks like MMMU (55.5 vs. 59.0) and MMVet (67.8 vs. 73.5). In contrast, our framework achieves a superior Pareto frontier: we maintain highly competitive foundational perception—surpassing Revisual-R1~\cite{chen2025advancing} on MMVet and MMMU, while delivering state-of-the-art reasoning performance. This demonstrates that our method successfully resolves the strategic perception deficit, equipping the model with the requisite visual fidelity to support its advanced reasoning engine without sacrificing logical robustness.


\begin{table}[!htb] 
\centering
\caption{Performance on multimodal perception benchmarks.}
  \setlength{\tabcolsep}{1.5pt}
  \renewcommand{\arraystretch}{1.1}
  \small
  \setlength{\extrarowheight}{2pt} 
\begin{tabular}{l|c|c|c|c|c|c} 
\thickhline
Method &Hall. & MME & MMB &MMStar &MMMU &MMVet \\
\hline
Qwen2.5-VL~\cite{bai2025qwen2}  & 65.0 & 2180 & 82.5 & 63.0 & 52.5 & 67.1 \\
ThinkLite-VL~\cite{wang2025sota}  & \textbf{70.7} & \textbf{2378} & \textbf{85.5} & \textbf{67.0} &55.5 &67.8\\
Revisual-R1~\cite{chen2025advancing} & 69.2 & 2284 & 82.1 & \textbf{67.0} &50.6 & 49.8 \\
\hline
Ours & 68.7 & 2307  &82.0 &66.0 &\textbf{59.0} &\textbf{73.5}\\
\thickhline
\end{tabular}
\label{tab:multimodal_perception}
\end{table}


\subsection{Cognitive Behavior Analysis}
\begin{figure}[h]
    \centering
    \includegraphics[width=1.0\linewidth]{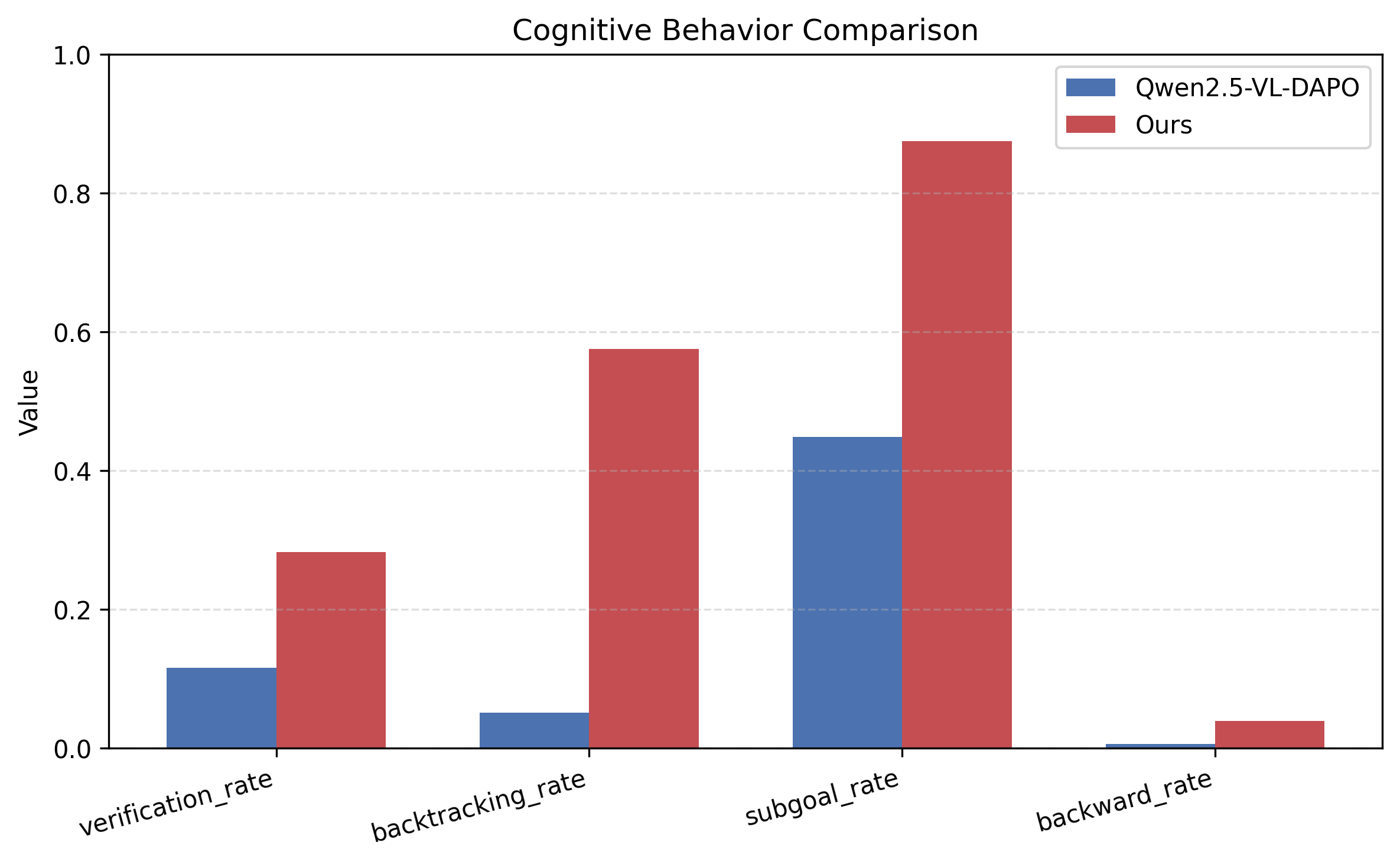} 
    \caption{Cognitive behavior comparison.}
    \label{fig:cognitive_behavior}
\end{figure}

Our framework is founded on the hypothesis that the two-stage cold-start curriculum cultivates essential cognitive behaviors that serve as prerequisites for effective reinforcement learning. To empirically validate this claim, we quantitatively measure the emergence of these structured cognitive patterns within model-generated reasoning trajectories.
We adopt the four cognitive behaviors proposed by Gandhi~\etal\cite{gandhi2025cognitive}: verfication, backtracking, subgoal setting and backward chaining.
We follow their classification pipeline using GPT-4o-mini to reliably identify these patterns in the outputs of the model.
We compute the Coginitive Behavior Emergence Ratio, which is the ratio between the number of emergence of visual behaviors and the number of samples.
We compare our full RL model against the Qwen2.5-VL-DAPO baseline on MathVision (mini)~\cite{wang2024measuring}, a benchmark specifically curated for evaluating long-chain mathematical reasoning. 
The results, depicted in Figure~\ref{fig:cognitive_behavior}, reveal a pronounced gap in behavioral emergence: our method exhibits substantially higher frequencies of all four cognitive behaviors, with verification, backtracking and subgoal setting showing particularly dramatic improvements. 
By first building abstract reasoning templates through text-only SFT and then anchoring them to visual evidence via PG-CoT, our curriculum equips the model with the foundational `System 2' backbone necessary for strategic self-improvement. Consequently, when RL is applied, it refines these latent behaviors rather than attempting to learn them from scratch, enabling the model to dynamically schedule perception at genuine moments of uncertainty. 
The enhanced cognitive flexibility translates directly into superior reasoning performance, as evidenced by our state-of-the-art results on multimodal benchmarks in Table~\ref{tab:multimodal_perception}.

\subsection{Ablation Studies}
To validate the contributions of the key components within our proposed framework, we conduct a series of ablation studies. We evaluate the effects of our disentangled cold-start curriculum, perception grounded CoT data and pivotal perception reward. All evaluation results are performed with a maximum generation limit of 8k new tokens.

\subsubsection{The Effect of Disentangled Cold Start SFT}
Our multi-stage cold-start strategy, which first text-only SFT to build robust cognitive behaviors and then perception-grounded SFT to anchor reasoning to visual evidence, delivers superior overall performance compared to single-stage alternatives. 
As shown in Table~\ref{tab:ablation_cold_start}, our two-stage curriculum achieves the highest average score (50.8) across benchmarks, demonstrating the efficacy of disentangled training.
Text-Only SFT excels on reasoning-intensive tasks like MathVision (44.4) and MathVerse (50.3), confirming that abstract logical priors benefit complex inference, yet it catastrophically underperforms on WeMath (34.6), where visual grounding is paramount. Conversely, Standard Multimodal SFT achieves the highest WeMath score (47.3) but suffers on reasoning-heavy benchmarks (33.3 on MathVision, 40.0 on MathVerse), where conflated objectives hinder systematic thinking. This performance dichotomy precisely validates our hypothesis: entangled training forces models to compromise between reasoning and perception, resulting in specialization rather than generalization.

\begin{table}[h]
  \centering
  \caption{Ablations on disentangled SFT curriculum. Text-Only: cognitive warm-up. Multimodal-Only: standard multimodal SFT. Two-Stage: our full curriculum.}
  \setlength{\tabcolsep}{2pt}
  \renewcommand{\arraystretch}{1.1}
  \small
  \setlength{\extrarowheight}{2pt}
  \scalebox{0.93}{
    \begin{tabular}{l|cccc|c}
      \thickhline
      Method & MathVista & MathVision & WeMath & MathVerse & Avg. \\
      \hline
      Text-Only &68.8 &\textbf{44.4} &34.6 &\textbf{50.3} & 49.5\\
      Multimodal-Only & 68.6 & 33.3 & \textbf{47.3} & 40.0 & 47.3 \\
      Two-Stage & \textbf{70.1} & 40.4 & 45.6 & 47.1 & \textbf{50.8} \\
      \thickhline
    \end{tabular}
    }
  \label{tab:ablation_cold_start}
\end{table}

\subsubsection{The Effect of Perception Grounded CoT Data}
We isolate the impact of our PG-CoT data paradigm by comparing training trajectories with and without perception grounding, both before and after RL. 
Table~\ref{tab:ablation_cot} reveals a compelling interaction: while PG-CoT data alone provides no average gain (both SFT variants achieve 50.8), it unlocks dramatically larger improvements under RL.

Introducing perception grounding during SFT initially degrades performance on MathVista (67.8 vs. 70.1) and MathVision (39.9 vs. 40.4), while providing slight boosts on WeMath and MathVerse. This confirms our hypothesis: teaching perceptual structure without strategic timing can hurt reasoning fluidity. However, when we apply our Pivotal Perception RL (PRL) to this PG-CoT foundation, performance leaps dramatically: +5.6 on MathVista, +4.3 on MathVision, +7.2 on WeMath, and +4.2 on MathVerse, achieving a 56.1 average (+5.3 over Original CoT + RL).
The key insight is that PG-CoT data serves as an enabling bridge rather than a direct performance booster. It provides the structural scaffolding for grounded reasoning, while PRL learns \textit{when} to invoke perception by coupling it to linguistic markers of cognitive uncertainty. 

\begin{table}[h]
  \centering
  \setlength{\tabcolsep}{2pt}
  \renewcommand{\arraystretch}{1.1}
  \small
  \setlength{\extrarowheight}{2pt}
  \caption{Ablations on perception grounded CoT dataset.}
  \scalebox{0.93}{
    \begin{tabular}{l|l|cccc|c}
      \thickhline
     Dataset & Stage & MathVista & MathVision & WeMath & MathVerse & Avg. \\
      \hline
      \multirow{2}{*}{Original CoT} &SFT & 70.1 & 40.4 & 45.6 & 47.1 &50.8 \\
      &+ RL & 72.7 & 40.7 & 51.6 & 50.1 & 53.8 \\
      \hline
      \multirow{2}{*}{PG-CoT} & SFT & 67.8 & 39.9 & 46.2 & 49.1 & 50.8 \\
       & + RL & \textbf{73.4} & \textbf{44.2} & \textbf{53.4} & \textbf{53.3} & \textbf{56.1} \\
      \thickhline
    \end{tabular}
  }
  \label{tab:ablation_cot}
\end{table}

\subsubsection{The Effect of Pivotal Perception Reward}
To isolate the contribution of our Pivotal Perception Reward, we compare RL variants with and without pivotal perception reward, all built upon the same Cold Start foundation. Table~\ref{tab:ablation_pivotal_reward} demonstrates that standard RL provides modest improvements, while our reward design unlocks substantial gains.
Standard RL rewards (\textit{w.o.} PR) yields limited improvements: +2.1 on MathVista, +0.3 on MathVision, +6.6 on WeMath, and +1.6 on MathVerse, averaging +2.6 points. While effective on certain benchmarks like WeMath, this approach fails to significantly enhance reasoning-heavy tasks such as MathVision, indicating that conventional rewards cannot effectively teach strategic perception timing.
Incorporating our Pivotal Perception Reward (\textit{w.} PR) drives dramatic performance leaps: +5.6 on MathVista, +4.3 on MathVision, +7.2 on WeMath, and +4.2 on MathVerse, achieving a 56.1 average (+5.3 over Cold Start). These gains are particularly pronounced on MathVision and MathVerse, benchmarks requiring fine-grained visual analysis.
This confirms that rewarding perception actions based on linguistic markers of cognitive uncertainty (metacognitive markers, structural transitions, re-examination cues) effectively teaches the model \textit{when} to perceive. 

\begin{table}[h]
  \centering
  \caption{Ablations on pivotal perception reward. All methods start from the same Cold Start checkpoint. Cold Start: Two-stage SFT with PG-CoT data. RL \textit{w.o.} PR: Standard RL with accuracy, format and length rewards only. RL \textit{w.} PR: Our full RL with Pivotal Perception Reward (PR).}
  \setlength{\tabcolsep}{2pt}
  \renewcommand{\arraystretch}{1.1}
  \setlength{\extrarowheight}{2pt}
  \small
    \begin{tabular}{l|cccc|c}
      \thickhline
      Method & MathVista & MathVision & WeMath & MathVerse & Avg. \\
      \hline
      Cold Start & 67.8 & 39.9 & 46.2 & 49.1 & 50.8 \\
      + RL \textit{w.o.} PR & 69.9 & 40.2 & 52.8 & 50.7 & 53.4 \\
      + RL \textit{w.} PR & \textbf{73.4} & \textbf{44.2} & \textbf{53.4} & \textbf{53.3} & \textbf{56.1} \\
      \thickhline
    \end{tabular}
  
  \label{tab:ablation_pivotal_reward}
\end{table}

\section{Conclusions}
In this work, we addressed the critical ``think longer, see less'' pathology in MLLMs, where progressive visual forgetting undermines long-chain reasoning. We argued this stems from entangled training, creating both a foundational cold-start deficiency (weak abstract reasoning) and a strategic perception deficit (no ``when-to-look'' policy).
To solve this, we proposed a framework to disentangle these skills. Our disentangled SFT curriculum first builds a robust text-only reasoning backbone before grounding it with novel Perception-Grounded CoT (PG-CoT) data. Subsequently, our Pivotal Perception Reward trains an RL agent to learn when to look by coupling perceptual actions to cognitive uncertainty. By decomposing the problem—learning to think, then to ground, and finally when to look—our work provides a curriculum for building more robust and strategic multimodal reasoners.

{\small \bibliographystyle{ieeenat_fullname} \bibliography{reference.bib}
}

\newpage
\onecolumn
\appendix
\section{More Implementation Details}
\label{sec: training details}
In the SFT stage, we utilized the LlamaFactory training framework~\cite{zheng2024llamafactory}. For text-only SFT, we used the GRAMMAR dataset~\cite{chen2025advancing}, which contains 47K samples focused on reasoning tasks. 
For the multimodal SFT phase, we collected and integrated open-source datasets to generate our proposed PG-COT dataset, such as ARES~\cite{chen2025ares} and CLEVR-Math~\cite{lindstrom2022clevr}. 


In the reinforcement learning stage, we employed the EasyR1 framework~\cite{zheng2025easyr1} and the Virl39K training dataset~\cite{wang2025vl}. The model was optimized using the DAPO algorithm~\cite{yu2025dapo} combined with our proposed composite reward function, and training was continued until the accuracy metric converged.
The specific training parameters for both SFT stages (which share the same configuration) and the RL stage are summarized in Table~\ref{tab:params}.
\begin{table}[h]
\centering
\caption{Key Hyperparameters for Training Stages}
\vspace{-2mm}
\label{tab:hyperparameters_side_by_side}
\scalebox{0.95}{
\begin{tabular}{ll|ll}
\toprule
\textbf{Component} & \textbf{Hyperparameter} & 
\textbf{Component} & \textbf{Hyperparameter} \\
\midrule
\multirow{8}{*}{\textbf{Cold Start}} & Learning Rate = \(5.0 \times 10^{-5}\) & 
\multirow{8}{*}{\textbf{DAPO}} & Max Response Length = 8192 \\
& Gradient Accumulation = 8 & & Rollout Batch Size = 512 \\
& Number of Epochs = 5 & & Actor Global Batch Size = 128 \\ 
& LR Scheduler = Cosine & & Learning Rate  = \(1.0 \times 10^{-6}\) \\
& Warmup Ratio = 0.05 & & Weight Decay = \(1.0 \times 10^{-2}\) \\
& Max Sequence Length = 32768 & & Optimizer Strategy = adamw \\
& Precision = BF16 & & Generation Temperature = 1.0 \\ 
& DeepSpeed = Zero2 & & Generation Top-p = 1.0 \\ 

\bottomrule
\end{tabular}
}
\label{tab:params}
\end{table}

\section{Details for Pivotal Reasoning Keywords}
\label{sec: Pivotal Reasoning Keywords}
We propose a pivotal perception reward to train the model's policy on strategic perception timing, based on the hypothesis that visual grounding should be triggered by linguistic markers of cognitive uncertainty. 
The reasoning-pivotal keywords in $\mathcal{K}$ are:
\begin{figure}[H]
\begin{keywordbox}{Pivotal Reasoning Keywords}
\textbf{Metacognitive Markers:}\\
\texttt{let me recalculate, let's revise, let me reconsider, let's rethink, let me check again, double-check, verify, re-evaluate, reassess, revisit, let's try again, wait, hold on, however, but, although, on second thought, wait a second, is that right, did I miss something, I'm not sure, I am not sure, I wonder, perhaps, maybe, I think, mistake, error, incorrect, wrong, I was mistaken, that's not right, this contradicts, I need to reconsider, doesn't make sense, I made an error, contradiction, flaw, invalid}\\[0.5em]
\textbf{Structural Transitions:}\\
\texttt{first, second, third, fourth, fifth, sixth, seventh, eighth, ninth, tenth, step, steps, step 1, step 2, option, options, approach, approaches, method, methods, way, ways, process, procedure, stage, stages, phase, phases, part, parts, section, sections, aspect, aspects, next, then, after, following, subsequently, initially, finally, lastly, ultimately, sequence, order, progression}\\[0.5em]
\textbf{Visual Re-Examination Cues:}\\
\texttt{observe, observing, observation, observations, notice, noticing, noticed, see, seeing, saw, seen, look, looking, looked, examine, examining, examined, examination, check, checking, checked, identify, identifying, identified, identification, recognize, recognizing, recognized, recognition, detect, detecting, detected, detection, find, finding, found, discover, discovering, discovered, discovery, analyze, analyzing, analyzed, analysis, consider, considering, considered, consideration, assess, assessing, assessed, assessment, inspect, inspecting, inspected, inspection, review, reviewing, reviewed, investigate, investigating, investigated, investigation, explore, exploring, explored, exploration}
\end{keywordbox}
\vspace{-2mm}
\label{fig: keywords}
\end{figure}

\section{Details for Perception Grounded CoT Data Generation}
\label{sec: data generation}
We transform existing multimodal reasoning datasets into PG-CoT format using a teacher MLLM, \ie Seed1.5-VL~\cite{guo2025seed1}. 
The data generation pipeline operates in three steps: (1) analyze the original CoT to identify logical breakpoints where visual evidence is required, (2) generate fine-grained observations anchored to specific visual elements, and (3) insert perception segments while preserving reasoning fidelity.
The prompts for these three steps are as follows:

\begin{promptbox}{1. Prompt for Image Description Generation}
\begin{verbatim}
Please provide an extremely detailed and comprehensive description of this image in English.

Requirements:
   1. Describe ALL visible text content(including titles, body text, numbers, formulas, labels, etc.)
  2. Describe ALL visual elements (shapes, charts, symbols, photos, illustrations, etc.),
     including their colors, positions, and size relationships
  3. Describe the spatial layout and arrangement of elements
  4. If there are tables, charts, or other structured information, describe their content
     in detail
  5. Be as thorough as possible so that a reader could understand the complete content of
     the image based solely on your description

Please start describing directly without using numbered points or specific formatting. 
Use natural, flowing language to present all the information in the image comprehensively.
\end{verbatim}
\end{promptbox}

\newpage
\begin{promptbox}{2. Prompt for Perception Location Recommendation}
\begin{verbatim}
You are an expert in multimodal reasoning and data annotation. 
Your task is to review a text-only Chain-of-Thought (CoT) reasoning process for a question that 
originally included an image. You need to:
  1. Suggest where perception segments (i.e., visual observations) could be inserted in the 
     reasoning, and specify what type of visual information should be described in each segment.
   2. Identify any sentences or phrases within the reasoning that already describe visual information, 
     and recommend wrapping these with <perception>...</perception> tags.

---
Original Question:
{Insert the original question here}

Thinking Content (MUST be preserved exactly as shown below):
{Insert the original CoT reasoning here}

Important Context:
The original question was accompanied by an image. The reasoning provided here is based only on
text, but in a multimodal setting, certain reasoning steps could be enhanced by directly grounding
them in visual evidence.

---
Core Instructions:
  1. Carefully read the entire CoT reasoning to understand its logic and structure.
  2. For each reasoning step, ask: "Would this step benefit from being grounded in visual
  evidence from the image?"
  3. Suggest specific points (only after punctuation marks) where a perception segment could be
  inserted to support or guide the reasoning.
   4. For each suggested perception segment, clearly specify what type of visual information should
  be described (e.g., chart data, labels, object attributes, spatial relationships,
  mathematical expressions, symbols, text, numbers, geometric shapes, etc.).
  5. Carefully examine the reasoning content and identify any sentences or phrases that already
  directly describe visual information from the image. For these, recommend wrapping them with
  <perception>...</perception> tags.
  6. Do NOT generate actual perception content—only indicate where and what should be described,
  and which existing segments should be wrapped, as if preparing instructions for a multimodal
  annotator.
  7. Your suggestions must be based entirely on the reasoning process and the original question.
  Do not invent or assume visual details that are not logically implied or necessary.
  8. Do NOT remove, rewrite, or summarize any part of the original reasoning. Your job is to annotate
  the reasoning process, not to improve or alter it.

---
Output Format:
- For each suggested insertion point, list:
  1. The exact location (e.g., after which sentence or phrase).
   2. The type of visual information that should be described in the perception segment (in clear, 
   concise language).
- For each existing sentence or phrase that describes visual information, specify its location and 
  recommend wrapping it with <perception>...</perception> tags.
- Maintain the order and structure of the original reasoning.
- When specifying the type of visual information, use the same language as the original content

Continued on next page ...
\end{verbatim}
\end{promptbox}

\begin{promptbox}{2. Prompt for Perception Location Recommendation (Cont.)}
\begin{verbatim}

---
Rules for Perception Segment Suggestions:
- Only suggest perception insertions after logical break points (punctuation marks, not in the
middle of sentences or formulas).
- The suggested visual information must be directly relevant to the reasoning step and necessary for
answering the main question.
- Do not suggest irrelevant or distracting visual observations.
- Distribute suggestions evenly throughout the reasoning trace, where they would add value.
- If a reasoning step does not require visual grounding, do not suggest a perception segment.
- Do not suggest insertions in the middle of mathematical formulas or equations.
- For existing visual description sentences, only recommend wrapping those that are directly and
factually describing the image.

---
Examples of CORRECT suggestions:
- After "We need to solve for the range."
  - Suggested perception: Describe the chart showing export data for four years.
- After "Looking at the chart data: 2004 shows $8.66 billion, 2005 shows $10.92 billion, 2006 shows
$13.41 billion, 2007 shows $16.7 billion."
  - Suggested perception: Describe the exact values and labels visible in the chart.
- Sentence: "The image shows a right triangle with sides labeled 3, 4, and 5."
  - Recommendation: Wrap this sentence with <perception>...</perception> tags.

Examples of INCORRECT suggestions:
- In the middle of a sentence or formula.
- After reasoning steps that do not rely on visual information.
- Suggesting perception segments that are not relevant to the question.
- Wrapping sentences that do not actually describe visual evidence.

---
Begin your output now:
\end{verbatim}
\end{promptbox}

\begin{promptbox}{3. Prompt for Perception Segment Insertion}
\begin{verbatim}
You are an expert in multimodal reasoning and data annotation. 
Your task is to enhance a text-only Chain-of-Thought (CoT) reasoning for a question that originally 
included an image. 
You will use the provided suggestions to:
  1. Wrap existing sentences or phrases that directly describe visual information
   with <perception>...</perception> tags, according to the suggestions.
  2. Insert new perception segments at the recommended locations, following the suggestions on
   what type of visual information to describe.
   3. Ensure all perception segments are concise, factually accurate, and based only on what can be
   directly observed in the image.
   4. For each perception segment, cross-check with the provided image description for consistency,
   but rely primarily on your own observation of the image.
  5. If the suggested visual evidence is not present in the image, simply skip inserting that
   perception segment.

---
Original Question:
{Insert the original question here}

Original Image:
{Insert the image here or provide the image file}

Image Description:
{Insert the pre-generated image description here}

Thinking Content (MUST be preserved exactly as shown below):
{Insert the original CoT reasoning here}

Perception Suggestions:
{Insert the list of suggestions, including:
  - Locations for inserting new perception segments and what to describe
  - Locations of existing visual description sentences/phrases to be wrapped}

---
Core Instructions:
  1. Carefully observe the provided image to understand all visual elements present, including:
  natural images, photographs, charts, graphs, diagrams, mathematical expressions, formulas,
  text content, numbers, symbols, geometric shapes, and any other visible elements.
  2. Read the provided Chain-of-Thought reasoning and the perception suggestions.
  3. For each suggestion to wrap an existing sentence or phrase, locate it in the reasoning and
  wrap it with <perception>...</perception> tags.
  4. For each suggestion to insert a new perception segment, add a <perception>...</perception>
  block at the specified location (only after punctuation), describing the recommended type of
  visual information observed in the image.
  5. For each perception segment, cross-check with the image description for consistency, but base
  your final judgment on your own observation.
  6. If the suggested visual evidence is not present in the image, simply skip inserting that
  perception segment.
  7. Each perception segment must contain a single, concise, factually accurate sentence that is
  directly verifiable from the image.

Continued on next page ...
\end{verbatim}
\end{promptbox}

\begin{promptbox}{3. Prompt for Perception Segment Insertion (Cont.)}
\begin{verbatim}
  8. Do not invent or assume details not visible in the image. Base all perception content on direct 
  observation.
  9. Do not remove, rewrite, or summarize any part of the original reasoning. Preserve the logical
  structure and meaning.
 10. Use the same language as the original reasoning when writing perception segments.
 11. Insert perception segments only at logical break points (after punctuation marks,
  never in 
  the middle of sentences or formulas).
  
---
Output Format:
- Return the enhanced reasoning content, with <perception>...</perception> tags 
correctly wrapped around existing visual description sentences/phrases, and new <perception> 
segments inserted at the recommended locations.
- Each <perception> segment should be concise and directly describe what is observable in the image.

---
Examples of CORRECT placement:
"<perception>The image shows a right triangle with legs of length 3 and 4</perception>
According to the Pythagorean theorem, we need to calculate the hypotenuse length."
"<perception>The bar chart shows sales data for three different products, with Product A as blue 
bars, Product B as red bars, and Product C as green bars</perception>We need to compare the sales 
performance of these three products."

---
Begin your output now:
\end{verbatim}
\end{promptbox}

\section{More Case Studies}
\label{sec: examples}
We provide three illustrative prediction examples, contrasting our approach with the Qwen2.5-VL-DAPO baseline. Examination of our reasoning trajectories reveals that perception segments are systematically invoked following pivotal reasoning keywords, validating the efficacy of our reward design in learning strategic perception timing.

\clearpage
\newpage
\begin{figure}[h]
    \centering
    \begin{vqaexample}[width=\textwidth]{Example 1}
        \begin{minipage}[t]{0.25\textwidth}
            \centering
            \includegraphics[width=\textwidth]{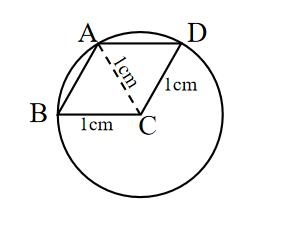}
        \end{minipage}
        \hfill
        \begin{minipage}[t]{0.65\textwidth}
            \begin{tcolorbox}[
                enhanced,
                colback=questionbg,
                colframe=bordercolor!70,
                boxrule=0.8pt,
                arc=3pt,
                left=6pt,
                right=6pt,
                top=4pt,
                bottom=4pt,
                fonttitle=\bfseries\color{white},
                title=Question: ,
                attach title to upper
            ]
                \ As shown in the figure, a circle is drawn with center C, and points A and B are exactly on the circle. The lengths of BC, AC, and CD are as shown in the figure, and AB = CD. What is the measure of $\angle$ ABC?\\ A. 45°; B. 60°; C. 72°; D. 90°; E. No correct answer
            \end{tcolorbox}
        \end{minipage}
        
        
\begin{tcolorbox}[
    enhanced,
    breakable,
    colback=answerbg2,
    colframe=bordercoloryellow!70,
    boxrule=0.8pt,
    arc=3pt,
    left=6pt,
    right=6pt,
    top=4pt,
    bottom=4pt,
    fonttitle=\bfseries\color{white},
    title=Answer (Qwen2.5-VL-DAPO): ,
    attach title to upper
]

    \textbf{<think>}

    To determine the measure of $\angle ABC$, we need to recognize that since AB = CD and all line segments are given as 1 cm, quadrilateral ABCD is a rhombus (as all sides are equal). Given that C is the center of the circle, triangle ABC and triangle ADC are isosceles triangles with AC = BC = CD = AD = 1 cm.

    Since AB = CD = 1 cm, quadrilateral ABCD is a rhombus. Because it lies inscribed in a circle (a cyclic quadrilateral), it must be a rectangle (as opposite angles sum to 180° and all sides are equal implying right angles at vertices A and C).

    Thus, $\angle ABC$ must be a right angle (90°) because in a cyclic quadrilateral where opposite sides are equal and inscribed in a circle, it forms a rectangle implying right angles at vertices.

    Therefore, $\angle ABC = 90°$.

    \textbf{</think>}

        \textbf{\textcolor{red}{\textbackslash \textbackslash boxed\{D\} (WRONG)}}
\end{tcolorbox}

\begin{tcolorbox}[
    enhanced,
    breakable,
    colback=answerbg,
    colframe=bordercolorgreen!70,
    boxrule=0.8pt,
    arc=3pt,
    left=6pt,
    right=6pt,
    top=4pt,
    bottom=4pt,
    fonttitle=\bfseries\color{white},
    title=Answer (Ours): ,
    attach title to upper
]

    \textbf{<think>}
    
    Okay, let's try to solve this problem. \textbf{\color{bordercolorblue}{First, let's look at the diagram.}} \textbf{\color{bordercolorgreen}{<perception>There's a circle with center C, so points A and B are on the circle, which means CA and CB are radii</perception>}}. The problem says BC, AC, and CD are as shown. The lengths: BC is 1 cm, AC is 1 cm, CD is 1 cm. Oh, so CA = CB = CD = 1 cm. So triangle ABC has CA = CB = 1 cm, and AB = CD. But CD is also 1 cm, so AB = 1 cm. Wait, so triangle ABC has sides CA = CB = AB = 1 cm. That makes triangle ABC equilateral. So all angles are 60 degrees. Therefore, angle ABC is 60°, which is option B.

    \textbf{\color{bordercolorblue}{Let me double-check.}} \textbf{\color{bordercolorgreen}{<perception>The diagram clearly labels CA, CB, and CD as 1cm each, and AB is equal in length to CD</perception>}} The center is C, so CA and CB are radii, so CA = CB = 1 cm. CD is also 1 cm, so AB = CD = 1 cm. So triangle ABC has three sides of 1 cm, so it's equilateral. Therefore, angle ABC is 60°, which is option B. \textbf{\color{bordercolorblue}{That seems right.}} \textbf{\color{bordercolorgreen}{<perception>The geometric arrangement confirms triangle ABC has all sides equal, identifying it as an equilateral triangle</perception>}} Maybe I'm missing something? Let me think again. The problem states AB = CD, and CD is 1 cm, so AB is 1 cm. CA and CB are both 1 cm, so triangle ABC is equilateral. So angle ABC is 60°, which is option B. So the answer should be B. Wait, but let me check the options again. Option B is 60°, so boxed B. So {\textbackslash \textbackslash boxed\{B\}}.
    
    \textbf{</think>}
    
    To solve the problem, we analyze the given diagram and information:

    \textbf{\color{bordercolorgreen}{<perception>The circle has center C, so CA, CB, and CD are radii. From the diagram, CA = CB = CD = 1 cm.</perception>}}

    AB is given to be equal to CD, so AB = 1 cm.

    In triangle ABC, CA = CB = AB = 1 cm, making triangle ABC equilateral.

    In an equilateral triangle, all angles are 60°, so $\angle ABC = 60°$.

    Thus, the measure of $\angle ABC$ is 60°, corresponding to option B.

    \textbf{\textcolor{Red}{\textbackslash \textbackslash boxed\{B\} (CORRECT)}}
\end{tcolorbox}

    \end{vqaexample}
    \label{fig:caption_geo_example}
\end{figure}

\clearpage
\newpage
\begin{figure}[!ht]
    \centering

\begin{vqaexample}[width=\textwidth]{Example 2}
    \begin{minipage}[t]{0.25\textwidth}
        \centering
        \includegraphics[width=\textwidth]{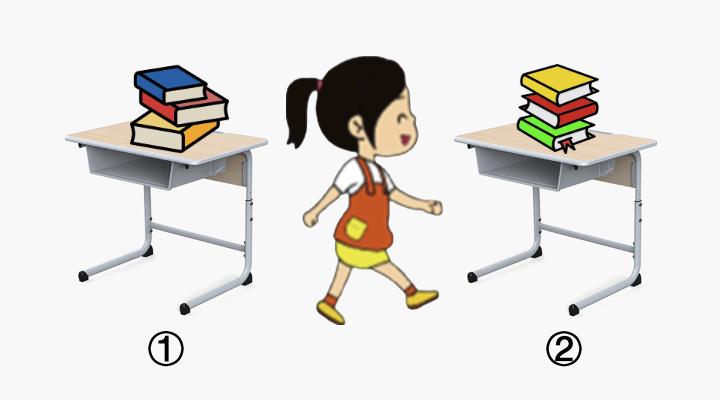}
    \end{minipage}
    \hfill
    \begin{minipage}[t]{0.65\textwidth}
        \begin{tcolorbox}[
            enhanced,
            colback=questionbg,
            colframe=bordercolor!70,
            boxrule=0.8pt,
            arc=3pt,
            left=6pt,
            right=6pt,
            top=4pt,
            bottom=4pt,
            fonttitle=\bfseries\color{titlecolor},
            title=Question: ,
            attach title to upper
        ]
            As shown in the diagram, the color of the book at the bottom of the desk in front of Lucy is ( ) \\ A. Red; B. Green; C. Blue; D. Yellow; E. No correct answer
        \end{tcolorbox}
    \end{minipage}

\begin{tcolorbox}[
    enhanced,
    breakable,
    colback=answerbg2,
    colframe=bordercoloryellow!70,
    boxrule=0.8pt,
    arc=3pt,
    left=6pt,
    right=6pt,
    top=4pt,
    bottom=4pt,
    fonttitle=\bfseries\color{white},
    title=Answer (Qwen2.5-VL-DAPO): ,
    attach title to upper
]

\textbf{<think>}
To determine the color of the book at the bottom of the desk in front of Lucy, I need to identify which desk is in front of her.Since the girl is positioned between the two desks, the desk on the left (labeled 1) is in front of her.

The books on desk 1 (from left to right) are:

Yellow

Red

Blue

Therefore, the book at the bottom of desk 1 is yellow.

Thus, the correct answer is D.
\textbf{</think>}

\textbf{\textcolor{red}{\textbackslash \textbackslash boxed\{D\} (WRONG)}}

\end{tcolorbox}

\begin{tcolorbox}[
    enhanced,
    breakable,
    colback=answerbg,
    colframe=bordercolorgreen!70,
    boxrule=0.8pt,
    arc=3pt,
    left=6pt,
    right=6pt,
    top=4pt,
    bottom=4pt,
    fonttitle=\bfseries\color{white},
    title=Answer (Ours): ,
    attach title to upper
]

\textbf{<think>}
Okay, let's try to figure this out. The question is about the color of the book at the bottom of the desk in front of Lucy. \textbf{\color{bordercolorblue}{First, I need to look at the diagram.}} \textbf{\color{bordercolorgreen}{{<perception>} The diagram has two desks: desk 1 on the left and desk 2 on the right. Lucy is between them. The desk in front of her is desk 2, because she's facing desk 2. \textbf{</perception>}}}
Now, let's check the books on desk 2. The books there are stacked: yellow, red, green? Wait, no, wait. Wait, the books on desk 2 are a stack. \textbf{\color{bordercolorblue}{Let me see:}} \textbf{\color{bordercolorgreen}{<perception>the top book is yellow, then red, then green</perception>}}. So the bottom book is green. Let me consider the options, the options are A. Red, B. Green, C. Blue, D. Yellow. So the bottom book is green, which is option B. \textbf{\color{bordercolorblue}{But I need to confirm the picture,}} \textbf{\color{bordercolorgreen}{<perception>The image shows desk 2 with a stack of books where the bottommost book is green</perception>}} So the answer should be B. Wait, but wait, maybe I'm misunderstanding. Wait, the desk in front of Lucy—since Lucy is walking towards desk 2, the desk in front of her is desk 2. So the books on desk 2: the bottom one is green. So option B. That's correct. \textbf{\color{bordercolorblue}{Wait, but let me make sure.}} \textbf{\color{bordercolorgreen}{<perception>Both desks have stacks of books, and desk 2's bottom book is green as observed in the image</perception>}} So the answer is B. So boxed B. The options are A to E. So yes, B.
\textbf{</think>}

To determine the color of the book at the bottom of the desk in front of Lucy, we first identify which desk is in front of her. \textbf{\color{bordercolorgreen}{<perception>Lucy is positioned between two desks, and the desk facing her is desk 2. On desk 2, the books are stacked with the bottom book being green. </perception>}}

The options are A. Red, B. Green, C. Blue, D. Yellow. The correct answer corresponds to the green book at the bottom.

    \textbf{\textcolor{Red}{\textbackslash \textbackslash boxed\{B\} (CORRECT)}}

\end{tcolorbox}

\end{vqaexample}
    \label{fig:caption_geo_example}
\end{figure}

\clearpage
\newpage
\begin{figure}[!ht]
    \centering
\begin{vqaexample}[width=\textwidth]{Example 3}
    \begin{minipage}[t]{0.25\textwidth}
        \centering
        \includegraphics[width=\textwidth]{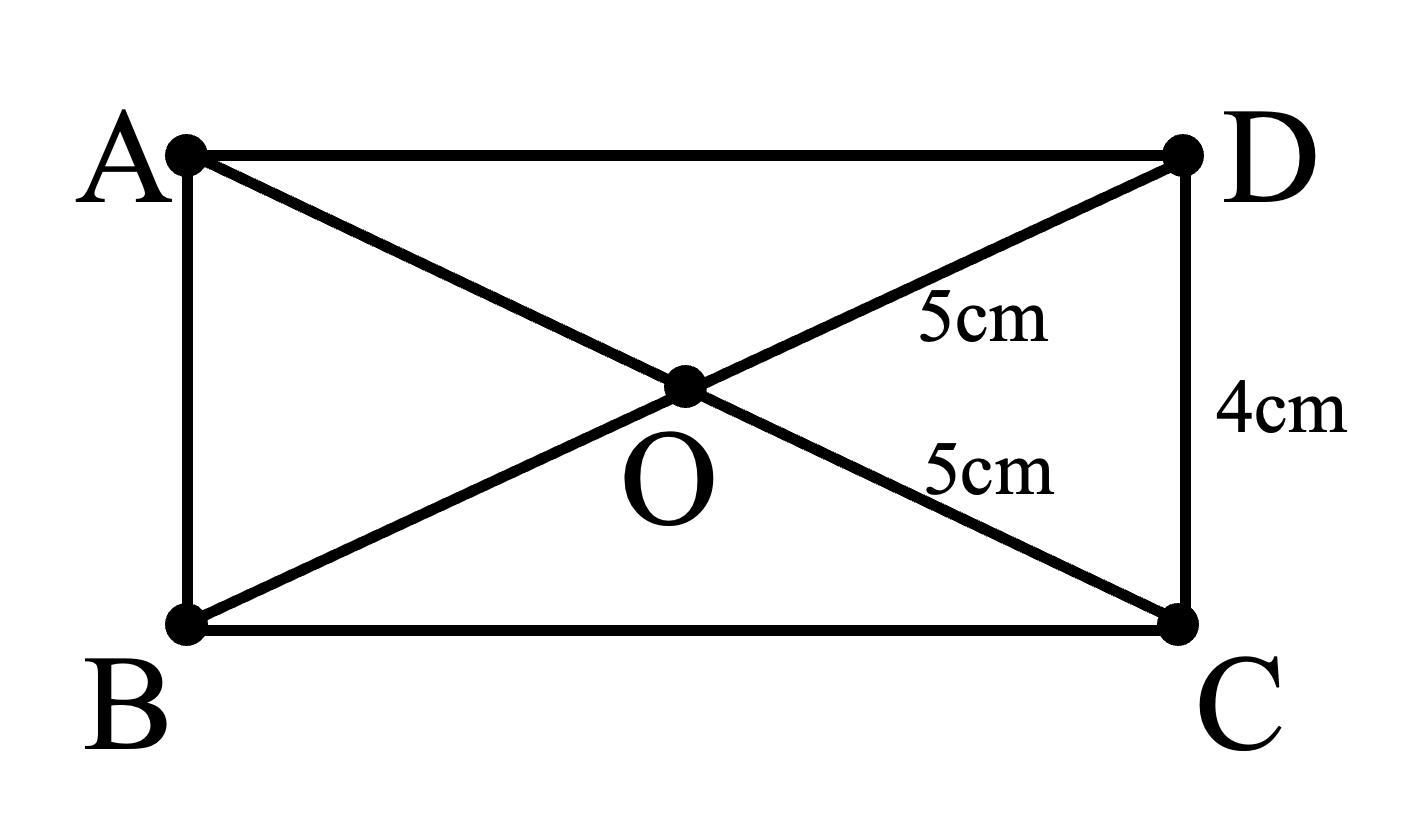}
    \end{minipage}
    \hfill
    \begin{minipage}[t]{0.65\textwidth}
        \begin{tcolorbox}[
            enhanced,
            colback=questionbg,
            colframe=bordercolor!70,
            boxrule=0.8pt,
            arc=3pt,
            left=6pt,
            right=6pt,
            top=4pt,
            bottom=4pt,
            fonttitle=\bfseries\color{titlecolor},
            title=Question: ,
            attach title to upper
        ]
            As shown in the figure, the diagonals of rectangle ABCD are AC and BD. What is the length of OC? \\( ) cmA. 3; B. 4; C. 5; D. 6; E. No correct answer
        \end{tcolorbox}
    \end{minipage}
    
\begin{tcolorbox}[
    enhanced,
    breakable,
    colback=answerbg2,
    colframe=bordercoloryellow!70,
    boxrule=0.8pt,
    arc=3pt,
    left=6pt,
    right=6pt,
    top=4pt,
    bottom=4pt,
    fonttitle=\bfseries\color{white},
    title=Answer (Qwen2.5-VL-DAPO): ,
    attach title to upper
]

    \textbf{<think>} 
    
    To determine the length of $OC$ in rectangle $ABCD$, we start by noting that the diagonals of a rectangle are equal and bisect each other. Therefore, $AC = BD$ and $AO = OC = BO = OD$. Given: $BD = 5$ cm.  Since BD is bisected at point O, it follows that:  $BO = OD = \frac{BD}{2} = \frac{5}{2} = 2.5 \text{ cm}$. Thus, $OC = 2.5 \text{ cm}$.  However, we need to find $OC$ in terms of the full length of diagonal components: Since $AC = BD = 5$ cm and $OC = AO = \frac{AC}{2}$.  Now, using the Pythagorean theorem in $\triangle DOC: AC^2 = AD^2 + DC^2$. Substituting the given lengths: $AC = \sqrt{AD^2 + DC^2} = \sqrt{4^2 + 5^2} = \sqrt{16 + 25} = \sqrt{41}$.  But since $OC$ is half of diagonal: $OC = \frac{AC}{2} = \frac{5}{2} = 2.5$  Thus, $OC = 2.5$  Therefore, the length of $OC$ is $2.5$.  
    
    \textbf{</think>} 
    
    \textbf{\textcolor{red}{\textbackslash \textbackslash boxed\{D\} (WRONG)}}
    
\end{tcolorbox}

\begin{tcolorbox}[
    enhanced,
    breakable,
    colback=answerbg,
    colframe=bordercolorgreen!70,
    boxrule=0.8pt,
    arc=3pt,
    left=6pt,
    right=6pt,
    top=4pt,
    bottom=4pt,
    fonttitle=\bfseries\color{white},
    title=Answer (Ours): ,
    attach title to upper
]

    \textbf{<think>}
    
    Okay, let's see. The problem is about a rectangle ABCD with diagonals AC and BD intersecting at point O. We need to find the length of OC. First, I remember that in a rectangle, the diagonals are equal and they bisect each other. So, AC = BD, and O is the midpoint of both diagonals. That means AO = OC = BO = OD.  \textbf{\color{bordercolorblue}{Wait,}} \textbf{\color{bordercolorgreen}{<perception>the diagram shows that one of the segments is 5 cm</perception>}}. Wait, which one? Oh, the diagonal BD is labeled 5 cm? Wait, no, the segment OD is labeled 5 cm? \textbf{\color{bordercolorblue}{Wait, looking at the diagram:}} \textbf{\color{bordercolorgreen}{<perception>the diagonal BD is split into BO and OD</perception>}}, so if O is the midpoint, then BD = 2 * OC. Wait, but the problem says ``the diagonals of rectangle ABCD are AC and BD''. So AC and BD are the diagonals, and they intersect at O, so O is the midpoint. So OC is half of AC, but also, since diagonals are equal, AC = BD. \textbf{\color{bordercolorblue}{Let me check again.}} \textbf{\color{bordercolorgreen}{<perception>The image shows the segment OD labeled ``5cm''</perception>}} Since in a rectangle, diagonals bisect each other, so OC = OD. Therefore, OC is 5 cm. So the answer is C. \textbf{\color{bordercolorblue}{But wait, let me make sure.}} \textbf{\color{bordercolorgreen}{<perception>The geometric figure is a rectangle, and the diagonals of the rectangle clearly bisect each other visually.</perception>}} So yes, OC is equal to OD, which is 5 cm. So the answer is C. Then, let me check the options. The options are A.3, B.4, C.5, D.6, E. So the answer is C. So boxed C. \textbf{\color{bordercolorblue}{Wait, but wait, maybe I'm missing something.}} \textbf{\color{bordercolorgreen}{<perception>The diagram clearly marks the length of OD as 5 cm</perception>}} The problem says ``the length of OC''. Since O is the midpoint, OC is half of diagonal BD. BD is 2*OC, but OD is 5 cm, so OC is 5 cm. That's correct. So the answer is C.
    
    \textbf{</think>}

    To solve the problem, we recall that in a rectangle, the diagonals are equal and bisect each other. This means the diagonals \( AC \) and \( BD \) intersect at point \( O \), making \( O \) the midpoint of both diagonals. Thus, \( OC = OD \).\textbf{\color{bordercolorgreen}{<perception>From the diagram, the segment OD is labeled \( 5 \, \text{cm} \)</perception>}}. Since \( O \) is the midpoint, \( OC \) is equal to \( OD \). Therefore, the length of \( OC \) is \( 5 \, \text{cm} \).

        \textbf{\textcolor{Red}{\textbackslash \textbackslash boxed\{C\} (CORRECT)}}

\end{tcolorbox}
\end{vqaexample}

    \label{fig:caption_geo_example}
\end{figure}

\end{document}